\documentclass[letterpaper, 10 pt, conference]{IEEEtran}  

\IEEEoverridecommandlockouts                              

\usepackage{graphicx}
\usepackage{color}
\usepackage{cite}
\usepackage[hidelinks,bookmarks=false]{hyperref}
\hypersetup{
colorlinks=true,
linkcolor=red,
citecolor=green,
filecolor=magenta,
urlcolor=blue
}

\usepackage{multirow}
\usepackage{enumerate}
\usepackage{bm}
\usepackage{booktabs}
\usepackage{blindtext}
\usepackage[T1]{fontenc}
\usepackage{mathptmx} 
\DeclareMathAlphabet{\mathcal}{OMS}{cmsy}{m}{n}
\usepackage{times} 
\usepackage{amsmath} 
\usepackage{amssymb}  
\usepackage[ruled,linesnumbered]{algorithm2e}

\usepackage{xcolor}

\usepackage{bbm}
\usepackage{booktabs}
\usepackage{upgreek}
\usepackage{mathtools,xparse}
\usepackage{colortbl}

\newcommand{\PreserveBackslash}[1]{\let\temp=\\#1\let\\=\temp}
\newcolumntype{C}[1]{>{\PreserveBackslash\centering}p{#1}}
\newcolumntype{R}[1]{>{\PreserveBackslash\raggedleft}p{#1}}
\newcolumntype{L}[1]{>{\PreserveBackslash\raggedright}p{#1}}

\DeclarePairedDelimiter{\norm}{\lVert}{\rVert}
\NewDocumentCommand{\normL}{ s O{} m }{%
  \IfBooleanTF{#1}{\norm*{#3}}{\norm[#2]{#3}}_{L_2($\Omega$)}%
}

\DeclareFontFamily{U} {cmmi}{}

\DeclareFontShape{U}{cmmi}{m}{n}{
  <-6> cmmi5
  <6-7> cmmi6
  <7-8> cmmi7
  <8-9> cmmi8
  <9-10> cmmi9
  <10-12> cmmi10
  <12-> cmmi12}{}

\DeclareSymbolFont{Xcmmi} {U} {cmmi}{m}{n}

\DeclareMathSymbol{\psi}{\mathord}{Xcmmi}{32}

\title{DQ-GAT: Towards Safe and Efficient Autonomous Driving with Deep Q-Learning and\\ Graph Attention Networks}

\author{Peide~Cai,
        Hengli~Wang,
        Yuxiang~Sun,~\IEEEmembership{Member,~IEEE}
        and~Ming~Liu,~\IEEEmembership{Senior Member,~IEEE}
\thanks{This work was supported by Zhongshan Science and Technology Bureau Fund, under project 2020AG002, Foshan-HKUST Project no. FSUST20-SHCIRI06C, and Guangdong Basic and Applied Basic Research Foundation project no. 2020A0505090008, awarded to Prof. Ming Liu. \textit{(Corresponding author: Ming Liu.)}}
\thanks{Peide Cai and Hengli Wang are with the Department of Electronic and Computer Engineering, The Hong Kong University of Science and Technology, Hong Kong SAR, China (email: pcaiaa@connect.ust.hk, hwangdf@connect.ust.hk).}
\thanks{Ming Liu is with the Thrust of Robotics \& Autonomous Systems, The Hong Kong University of Science and Technology (Guangzhou), Nansha, Guangzhou, 511400, Guangdong, China, with the Department of ECE, The Hong Kong University of Science and Technology, Hong Kong SAR, China, and also with HKUST Shenzhen-Hong Kong Collaborative Innovation Research Institute, Futian, Shenzhen. (email: eelium@ust.hk).}
\thanks{Yuxiang Sun is with the Department of Mechanical Engineering, The Hong Kong Polytechnic University, Hong Kong SAR, China (email: yx.sun@polyu.edu.hk, sun.yuxiang@outlook.com).}
}
\begin{document}

\maketitle

\begin{abstract}
Autonomous driving in multi-agent dynamic traffic scenarios is challenging: the behaviors of road users are uncertain and are hard to model explicitly, and the ego-vehicle should apply complicated negotiation skills with them, such as yielding, merging and taking turns, to achieve both safe and efficient driving in various settings. Traditional planning methods are largely rule-based and scale poorly in these complex dynamic scenarios, often leading to reactive or even overly conservative behaviors. Therefore, they require tedious human efforts to maintain workability. Recently, deep learning-based methods have shown promising results with better generalization capability but less hand engineering efforts. However, they are either implemented with supervised imitation learning (IL), which suffers from dataset bias and distribution mismatch issues, or are trained with deep reinforcement learning (DRL) but focus on one specific traffic scenario. In this work, we propose DQ-GAT to achieve scalable and proactive autonomous driving, where graph attention-based networks are used to implicitly model interactions, and deep Q-learning is employed to train the network end-to-end in an unsupervised manner. Extensive experiments in a high-fidelity driving simulator show that our method achieves higher success rates than previous learning-based methods and a traditional rule-based method, and better trades off safety and efficiency in both seen and unseen scenarios. Moreover, qualitative results on a trajectory dataset indicate that our learned policy can be transferred to the real world for practical applications with real-time speeds. Demonstration videos are available at \url{https://caipeide.github.io/dq-gat/}.

\end{abstract}


\section{Introduction}

\IEEEPARstart{A}{utonomous} driving (AD) technology has made substantial progress in the last decade, moving from academic research\cite{urmson2008autonomous} to industrial practice\cite{kuutti2021survey}. Nevertheless, reliable and robust autonomous driving in urban areas remains an important challenge, primarily due to the following reasons: 1) There are diverse road topologies and structures (e.g., roundabouts, multi-lane streets, and intersections) with different traffic densities to consider\cite{Chen2019DeepIL}; 2) The complex and coupled interactions among multiple road agents are hard to model explicitly\cite{Chen2020RobotNI}; 3) The agent vehicle needs to intelligently make decisions in these uncertain scenarios to properly balance two contradictory driving objectives: safety (collision avoidance) and efficiency (time to goal).

\begin{figure}[t]
        \centering
        \setlength{\abovecaptionskip}{-0.1pt}
        \includegraphics[width = 0.9\columnwidth]{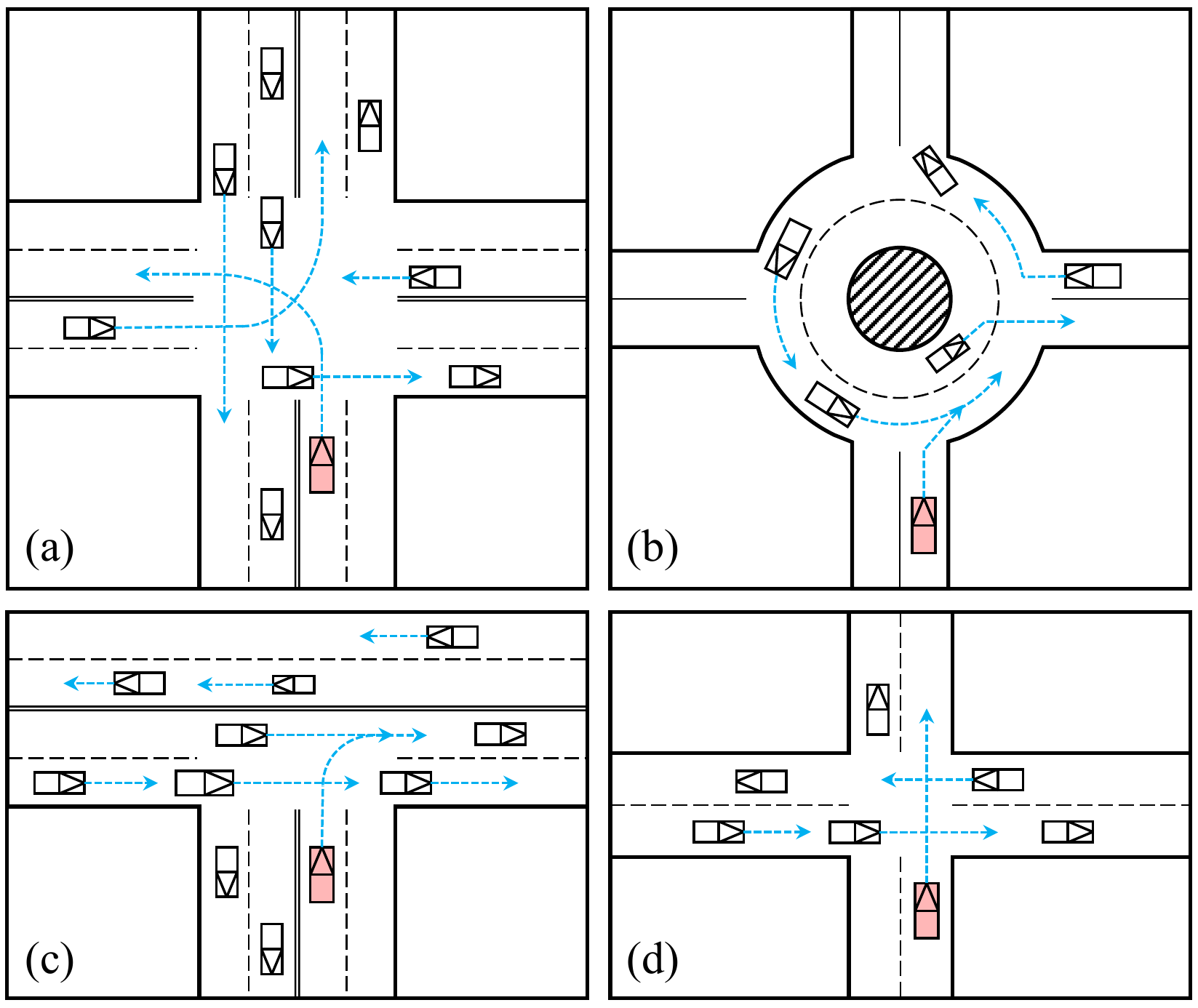}
        \caption{Sample unsignalized driving scenarios considered in this work: (a) unprotected left turns, (b) roundabouts, (c) merging, and (d) crossing. Rule-based methods have difficulties in tuning for complex interactions, often leading to reactive or even overly conservative behaviors. In this work we aim to train DRL-based agents that can act proactively amidst other vehicles to better trade off safety and efficiency of autonomous driving. The red and white boxes denote the ego and other vehicles, respectively.}
        \label{fig:motivation}
        \vspace{-0.4cm}
\end{figure}

Traditional planning methods are mainly based on hand-engineered heuristics \cite{urmson2008autonomous}, such as finite state machine (FSM)\cite{montemerlo2008junior, paden2016survey}. However, they are usually designed for a narrow set of particular use-cases, and require extremely tedious human efforts to maintain a rule database so that safety can be ensured \cite{Zeng2019EndToEndIN}. Moreover, new problems may arise as the rules increase, for example, how to solve new situations without forgetting the old ones, and how to balance cost functions in countless hard-to-model scenarios with conflicting objectives (e.g., safety and efficiency). Therefore, rule-based methods often lead to unnatural driving behaviors, or they completely fail in unexpected edge cases\cite{cao2020reinforcement}. For example, autonomous vehicles may slow down and stop in highly interactive scenarios (as shown in Fig. \ref{fig:motivation}) to ensure safety\cite{Zhou2020SMARTSSM}, known as the \textit{freezing robot problem}. However, such an overly conservative solution also causes confusion to other road users and even leads to traffic accidents\footnote{In California in 2018, 86\% of autonomous vehicle crashes were caused by other cars, resulting from the conservatism of the autonomous vehicles \cite{calicrash}.}. Due to the these limitations, traditional rule-based methods are "\textit{not robust to a varied world}", even according to their own authors\cite{urmson2008autonomous}.

In recent years, as an alternative, deep learning has advanced AD technology to a great extent. The ability to \textit{learn and self-optimize} its behavior from data alleviates the laborious engineering maintenance required to model all foreseeable scenarios, making a deep driving model well suited to AD problems in high-dimensional, nonlinear and dynamic environments\cite{Cai2020ProbabilisticEV, Chen2019LearningBC, codevilla2018end, codevilla2019exploring, cai2020vtgnet, liu2021gaze}. Most of these are implemented with supervised imitation learning (IL), which can efficiently extract driving knowledge from human demonstrations. However, this approach suffers from dataset bias\cite{codevilla2019exploring} and distribution mismatch (a.k.a., \textit{covariate shift})\cite{cai2021vision} problems. Another learning-based approach is deep reinforcement learning (DRL) \cite{Cai2020HighSpeedAD, cai2021vision, Kendall2019LearningTD, Leurent2019SocialAF, Chen2019ModelfreeDR}, where the agent proactively \textit{interacts} with the environment and learns knowledge from trial-and-error. However, current DRL-based methods have not been well designed for scalable AD in a uniform setup. Particularly, most work focuses on a specific network design tailored for scattered traffic scenarios such as single lanes\cite{Kendall2019LearningTD}, specific intersections\cite{Leurent2019SocialAF} and roundabouts\cite{Chen2019ModelfreeDR}, all with no \cite{Cai2020HighSpeedAD, cai2021vision} or low-level traffic dynamics, leaving it unclear if these models can generalize to more complex or unseen environments.

In this paper we propose DQ-GAT for autonomous driving in complex and dynamic scenarios. To avoid a model tailored for specific scenarios, we first design a graph attention-based network, which can process heterogeneous traffic information for generic driving scenarios. Afterwards, we design rewards and extend the model-free DRL method dueling double deep Q-learning (D3QN) into an asynchronous version. Finally, the network is trained at scale to adaptively balance safety and efficiency (i.e., \textit{ensuring driving safety while improving the efficiency as much as possible}) by proactively interacting with the environments.

This paper significantly extends our recent conference paper\cite{cai2021dignet}, and the improvements are multi-fold. First, we extend the original training pipeline from supervised learning to reinforcement learning, and demonstrate its superiority quantitatively in this work. Second, we conduct more thorough experiments with related works on DRL, IL and rule-based methods and provide more in-depth discussions. Moreover, we examine the zero-shot generalization performance and runtime speeds of our model in the real-world traffic dataset.

The main contributions are summarized as follows:

\begin{enumerate}

    \item We propose a novel graph attention-based network architecture to encode heterogeneous traffic information (i.e., road structures and vehicle states) and implicitly model inter-vehicle interactions in generic driving scenarios.
    
    \item We develop a parallel DRL framework to provide asynchronous and scalable training of the proposed network for autonomous driving without relying on supervisions.
    
    \item We conduct extensive evaluation in a high-fidelity driving simulator and show that our method balances safety and efficiency better than previous learning-based and rule-based methods, in both training and unseen scenarios.
    
    \item We show that our method can achieve sim-to-real policy transfer in an interaction-intensive real-world dataset, where real-time performance is also satisfied.
    
\end{enumerate}

\section{Related Work}

\subsection{Imitation Learning}

IL is the dominant approach for learning-based AD due to its sample efficiency, where a deep neural network is trained to mimic expert driving behaviors using supervised learning\cite{Cai2020ProbabilisticEV, Chen2019LearningBC, codevilla2018end, codevilla2019exploring, cai2020vtgnet, liu2021gaze}. Based on the collected demonstrations (i.e., observation-action pairs), the model compares the error between its prediction and the labelled data from human drivers, then adjusts its weights using gradient decent. 

\textbf{End-to-end driving.} With the powerful representation capabilities of deep neural networks, \textit{end-to-end} driving approaches \cite{codevilla2018end, Chen2019LearningBC, Cai2020ProbabilisticEV, codevilla2019exploring, cai2020vtgnet, liu2021gaze} directly take as input the raw sensor readings (e.g., LiDAR point clouds and camera images) to output control commands or future trajectories. For example, Codevilla \textit{et al.} \cite{codevilla2018end} proposed a conditional imitation learning approach that splits the network into multiple branches for discrete tasks such as \textit{follow lane} and \textit{turn left/right}. Follow-up works include \cite{Chen2019LearningBC} and \cite{cai2020vtgnet}. However, these methods cannot handle complex road topologies such as multi-lane streets or roundabouts. Recently, Cai \textit{et al.}\cite{Cai2020ProbabilisticEV} used global routes as direction to achieve robust end-to-end navigation in complex dynamic environments with multi-modal sensor fusion. However, as with previous methods, the learned policy is reactive without efficient interaction with other road users. 

To summarize, it is quite challenging to learn a direct mapping from high dimensional sensory observations to low dimensional motion plans, as the end-to-end approaches conflate two aspects of driving: \textit{learning to see} and \textit{learning to act}. Therefore, they suffer from the domain gap problem in the perception stage, which leads to poor generalization performance in new environments\cite{cai2021dignet}.

\textbf{Learning to drive by semantic abstraction.} Recently, another stream of work has arisen that uses \textit{semantic} information to learn driving policies (e.g., HD maps\cite{Chen2019DeepIL}, brid-eye-views (BEVs)\cite{bansal2019chauffeurnet}, and occupancy maps\cite{sadat2020perceive}). Compared to redundant sensory observations, this semantic information is a kind of concise and informative abstraction of perceptual results, and has better environmental consistency. These properties help the training process to focus on \textit{learning to act}, which is more efficient and generalizable. For example, the policy network of \cite{sadat2020perceive} takes as input hybrid features composed of the roadmap, traffic lights, route plans and dynamic objects to produce waypoints to follow. This has the advantage of helping the network to learn meaningful contextual cues behind the human driver's action and achieve more complex driving behaviors. However, \cite{sadat2020perceive} mainly shows its results on offline \textit{logged data}, and only performs closed-loop evaluation in simple environments with at most two obstacles. Such a problem also exists in other similar works\cite{bansal2019chauffeurnet}. According to \cite{Codevilla2018OnOE}, the driving performance can vary significantly between offline open-loop and online closed-loop tests, and the latter can better reveal the driving quality. In this work, the observation is similar to those of \cite{bansal2019chauffeurnet} and \cite{sadat2020perceive} but, differently, we focus more on dynamic, interactive and large-scale \textit{closed-loop} driving performance.

\textbf{Limitations.} Although the paradigm of IL is appealing, there still exist three major shortcomings that prevent IL from being applied to broader applications: 1) The cost of human driving data collection on a large scale can be prohibitive\cite{Leurent2019SocialAF}; 2) IL approaches are particularly sensitive to the \textit{dataset bias} issue, as different drivers, or even the same driver in different moods, might have different driving preferences, thus the learning objective might be dominated by the main modes in the training data\cite{codevilla2019exploring}; 3) IL performs well for states that are covered by the training distribution, but it generalizes poorly to new states due to compounding action errors\cite{Zhu2021ASO}, which is also referred to as \textit{distribution mismatch}.

\subsection{Deep Reinforcement Learning}
\label{subsec:intro_drl}

DRL is another popular training paradigm, where the agent interacts with the environment and gains knowledge through trial-and-error, aiming to maximize the sum of expected future rewards\cite{Cai2020HighSpeedAD, cai2021vision, Kendall2019LearningTD, Leurent2019SocialAF, Chen2019ModelfreeDR}. Therefore, it does not require expert labels and thus eliminates the dataset bias issue associated with IL. Furthermore, its online training paradigm also allows avoiding the distribution mismatch problem. Due to these advantages, DRL has shown promising results in various areas in decision making. For example, Wu \textit{et al.} \cite{wu2020reducing} proposed a triplet-average deep deterministic (TADD) policy gradient algorithm to reduce the estimation bias, and it achieves superior performance in the OpenAI gym environment. In addition, Dong \textit{et al.}\cite{dong2019dynamical} improved the performance of visual object tracking by dynamically optimizing its hyperparameters for changing sequences with deep Q-learning.

\textbf{Applications and limitations.} DRL has been applied to learn autonomous driving policies \cite{Cai2020HighSpeedAD, cai2021vision, Kendall2019LearningTD, Leurent2019SocialAF, Chen2019ModelfreeDR}. For example, \cite{Kendall2019LearningTD} trained an end-to-end policy for lane-following tasks on a slow vehicle, and \cite{Cai2020HighSpeedAD} achieved high-speed autonomous vehicle racing using model-free DRL methods. However, these works only consider static environments without interaction with other vehicles. By contrast, \cite{Bouton2020ReinforcementLW, Chen2019ModelfreeDR, Leurent2019SocialAF} and \cite{Saxena2020DrivingID} consider dynamic traffic scenarios and thus they are more applicable for urban driving. However, current methods have not been well designed for scalable AD in a uniform setup. Particularly, most works focus on specialized network design (e.g., input representations and reward) tailored for scattered traffic scenarios such as intersections\cite{Leurent2019SocialAF}, roundabouts\cite{Chen2019ModelfreeDR}, merging\cite{Bouton2020ReinforcementLW} and highway \cite{Kaushik2018OvertakingMI} scenarios, where one model cannot generalize across different scenarios due to the varied requirements of observation space. For example, \cite{Kaushik2018OvertakingMI} used a list of vehicle state vectors, such as position and velocity, to depict the environment. Although this representation is concise and precise, it lacks the information of driving contexts and would fail in complex scenarios such as multi-lane intersections. For example, in Fig. \ref{fig:motivation}, the vehicles should drive by obeying traffic rules according to the road structures. By contrast, processed BEVs\cite{Saxena2020DrivingID} are more suitable for context-aware driving, where all necessary information, including static road and dynamic vehicles, can be rasterized into pixels and jointly processed with convolution neural netowrks (CNNs). However, the drawback of this method is the information loss during rasterization. In this work, we combine the benefits of both pixel- and state-based methods to train agents that can drive in diverse urban scenarios.

\subsection{Graph Representation Learning}

Many real-world problems can be modeled with graphs where the nodes contain features of different entities, and edges represent interactions between entities. For example, in the field of video object segmentation, Lu \textit{et al.} used graphs to store frames as nodes and capture cross-frame correlations by edges\cite{lu2020video}. A challenge in learning on graphs is to find an effective way to get a meaningful aggregated feature representation to facilitate downstream tasks. Recently, graph neural networks (GNNs) have been shown to be effective in many applications such as social networks, personalized recommendation, video object segmentation\cite{lu2020video} and detection\cite{yin2021graph}. In this area, graph convolutional networks (GCNs) generalize the 2D convolution on grids to graph-structured data. When training a GCN, a fixed adjacency matrix is commonly adopted to aggregate the feature information of neighboring nodes. On the other hand, a graph attention network (GAT)\cite{velickovic2018graph} is a GCN variant that aggregates node information with weights learned in a self-attention mechanism. Such adaptiveness of GATs makes them commonly more effective than GCNs in graph representation learning.

\textbf{Applications and limitations.} Recently, graph neural networks (GNNs) have also been shown to be effective in robotics, such as in crowd robot navigation\cite{Chen2020RobotNI,Manso2019GraphNN}, where the robot can navigate safely in human crowds of various sizes. However, these works totally neglect the environmental structures, which is not particularly important for indoor robot navigation in restricted areas, but is non-negligible for outdoor driving problems, as introduced in Sec. \ref{subsec:intro_drl}. In this work, we borrow the idea of the GNN to model the traffic scenes as graphs, and use a context-aware GAT for autonomous driving in dense traffic. 

\begin{figure*}[t]
        \centering
        \setlength{\abovecaptionskip}{-0.1pt}
        \includegraphics[width = 2\columnwidth]{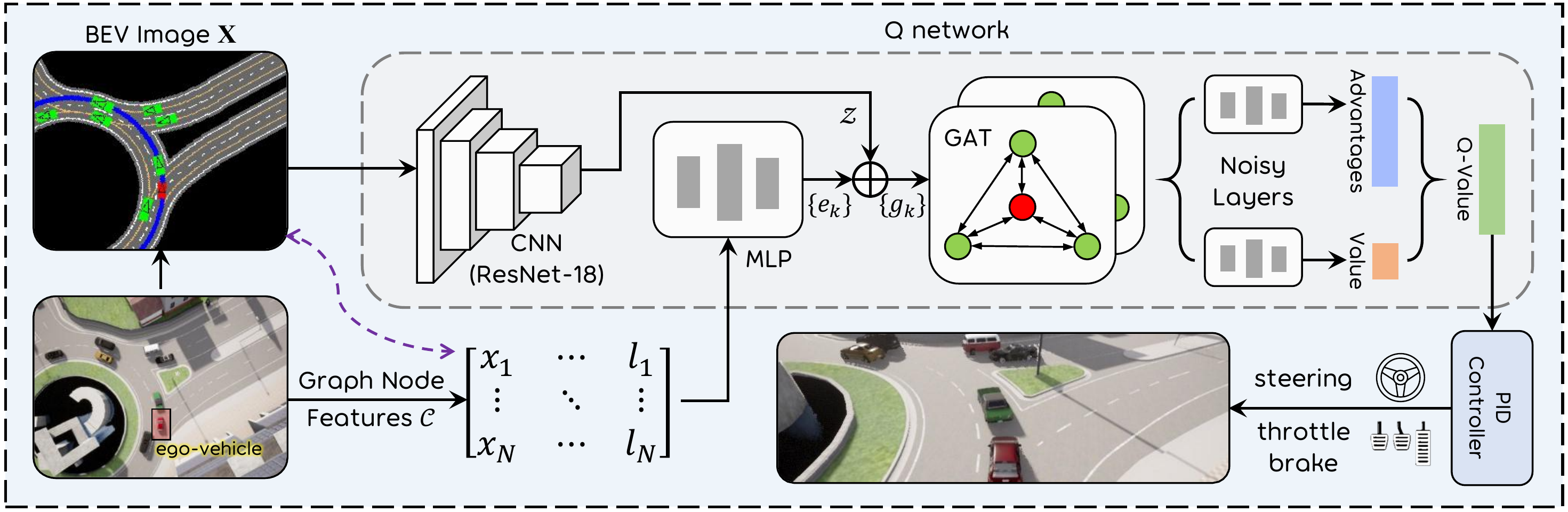}
        \caption{Schematic overview of the proposed method DQ-GAT for autonomous driving. We assume the input information is accessible with a functioning perception system or with a vehicle-to-everything (V2X) module, and focus on \textit{learning to act} of autonomous vehicles. The semantic BEV image and vehicle states (locations, velocities, etc.) are first encoded respectively by a CNN backbone (ResNet-18) and MLP layers. The derived feature vectors are then concatenated to construct a heterogenous scene-level graph $\{g_k\}$, which is further processed by a two-layer GAT and \textit{noisy} MLP layers (see Sec. \ref{subsubsec: noisy}) to compute Q-values for controlling the ego-vehicle.}
        \label{fig:architecture}
        \vspace{-0.4cm}
\end{figure*}

\section{Preliminaries}
In this paper, a value-based reinforcement learning algorithm called dueling double deep Q-learning (D3QN)\cite{wang2016dueling}, is applied to autonomous driving\footnote{In this work we choose to use value-based DRL as it has shown advanced intelligence beyond the level of human beings in certain areas like Go\cite{Silver2016MasteringTG}. In addition, it is also known to be more data efficient than the policy-based method, which is another family of DRL.}. In the following, we introduce the notation, terminology and algorithm for the system modeling and training.

\subsubsection{Markov Decision Process (MDP)}
The MDP process is the theoretical basis of reinforcement learning, and it can be formulated as a tuple containing five elements: $<\mathcal{S}, \mathcal{A}, R, f, \gamma>$, which denote the state space, action space, immediate reward, state transition model and the discount factor, respectively. Within this formulation, the agent interacts with environment and learns the policy $\pi$ by maximizing the expected discounted return $R_t = \sum_{\tau=t}^{\infty}\gamma^{\tau-t}r_{\tau}$, where $\gamma \in [0,1]$ trades-off the importance of immediate and future rewards. 

Given a policy $\pi$, the action-value (Q-value) of a state-action pair is defined as
\begin{equation}
Q_{\pi}(s_t, a_t)=\mathbb{E}\left[R_{t} \mid s_t, a_t, \pi\right],
\end{equation}
which can be computed using the Bellman equation:
\begin{equation}
Q_{\pi}\left(s_{t}, a_{t}\right)=\mathbb{E}[r_{t}+\gamma \mathbb{E}[Q_{\pi}\left(s_{t+1}, a_{t+1}\right)] \mid s_{t}, a_{t}, \pi].
\end{equation}
Finally, the optimal Q-value function can be written as
\begin{equation}
\label{eq:optimal_Q}
Q^{*}\left(s_{t}, a_{t}\right)=\mathbb{E}[r_t + \gamma \max_{a_{t+1}} Q^{*}\left(s_{t+1}, a_{t+1}\right) \mid s_{t}, a_{t}].
\end{equation}

\subsubsection{Double Deep Q-Learning}
It can be seen that once Eq. (\ref{eq:optimal_Q}) is computed, we can choose the optimal action $a_t^{*}$ with the largest Q-value to execute at state $s_t$. However, traditional tabular methods cannot scale to large state spaces, like images. In deep Q-learning, the $Q^{*}(s, a)$ in Eq. (\ref{eq:optimal_Q}) is approximated by a deep neural network $Q(s, a ; \theta)$ with parameters $\theta$. We use \textit{double} deep Q-learning\cite{van2016deep} to estimate this network, and optimize the following sequence of loss function at iteration $i$ based on the temporal-difference (TD) error:
\begin{equation}
\label{eq:td-loss}
L_{i}\left(\theta_{i}\right) =\mathbb{E}_{s, a, r, s^{\prime}}[\left(y_{i}-Q\left(s, a ; \theta_{i}\right)\right)^{2}],
\end{equation}
\begin{equation}
y_{i}=r+\gamma \hat{Q}(s^{\prime}, \underset{a^{\prime}}{\arg \max } Q\left(s^{\prime}, a^{\prime} ; \theta_{i}\right) ; \theta^{-}),
\end{equation}
where $\hat{Q}$ denotes the \textit{target network} with paramters $\theta^-$, which are updated by copying the weights of $Q(s,a;\theta)$ every $T$ gradient steps and are frozen in other intervals. In this work, ${Q}$ and $\hat{Q}$ share the same CNN encoder, and we set $T=1500$.

\subsubsection{Dueling Network Architecture} Based on the double DQN algorithm introduced above, Wang \textit{et al.}\cite{wang2016dueling} further proposed the \textit{dueling} network architecture for the Q network, named D3QN, where two streams of sub-networks are built to compute the state value $V_{\pi}(s)$ (scalar) and advantage functions $A_{\pi}(s, a)$ (vector of |$\mathcal{A}$|-dimensional) separately, as shown in Fig. \ref{fig:architecture}. These two branches are finally combined to compute the action values:
\begin{equation}
\begin{aligned}
Q(s, a ; &\theta)=V(s ; \theta)+(A(s, a ; \theta)-\frac{1}{|\mathcal{A}|} \sum_{a^{\prime}} A\left(s, a^{\prime} ; \theta\right)).
\label{eq:dueling}
\end{aligned}
\end{equation}

The decoupling operation of value and advantage in deep Q-networks has shown dramatic improvements in the challenging Atari gaming domain in terms of task performance and training speed. In this work, we extend this method into the area of autonomous driving, which will be methodologically introduced in the next section.

\subsubsection{Noisy Networks for Exploration}
\label{subsubsec: noisy}
Classical DRL methods use $\epsilon$-greedy strategies to randomly perturb the agent's policy and induce novel behaviors for exploration. However, these methods require hyperparameter tuning, and the induced \textit{local} dithering perturbations make it hard to generate diverse behaviors for efficient exploration. Therefore, we adopt the idea of \textit{noisy nets}\cite{fortunato2018noisy} for exploration in this work. This uses a noisy linear layer that combines a deterministic and noisy stream:
\begin{equation}
\bm{y}=(\bm{b}+\mathbf{W} \bm{x})+\left(\bm{b}_{\text {noisy }} \odot \epsilon^{b}+\left(\mathbf{W}_{\text {noisy }} \odot \epsilon^{w}\right) \bm{x}\right),
\end{equation}
where the parameters $\bm{b},\mathbf{W}, \bm{b}_{\text{noisy}}$ and $\mathbf{W}_{\text{noisy}}$ are learnable, while $\epsilon^{b}$ and $\epsilon^{w}$ are random variables. In this way, the amount of noise injected in the network is tuned automatically by the RL algorithm, allowing state-conditioned and self-annealing exploration. During testing, $\bm{b}_{\text{noisy}}$ and $\mathbf{W}_{\text{noisy}}$ are set to zero for stable policy deployment.

\section{Methodology}
\label{sec:method}
The structure of the proposed DQ-GAT for autonomous driving is shown in Fig. \ref{fig:architecture}. The goal is to drive safely and efficiently in complex and dynamic urban environments with different road layouts and other vehicles. 

\subsection{Training Scenarios}
\label{subsub:training_scenarios}
\begin{figure}[t]
        \centering
        \setlength{\abovecaptionskip}{-0.1pt}
        \includegraphics[width = 0.95\columnwidth]{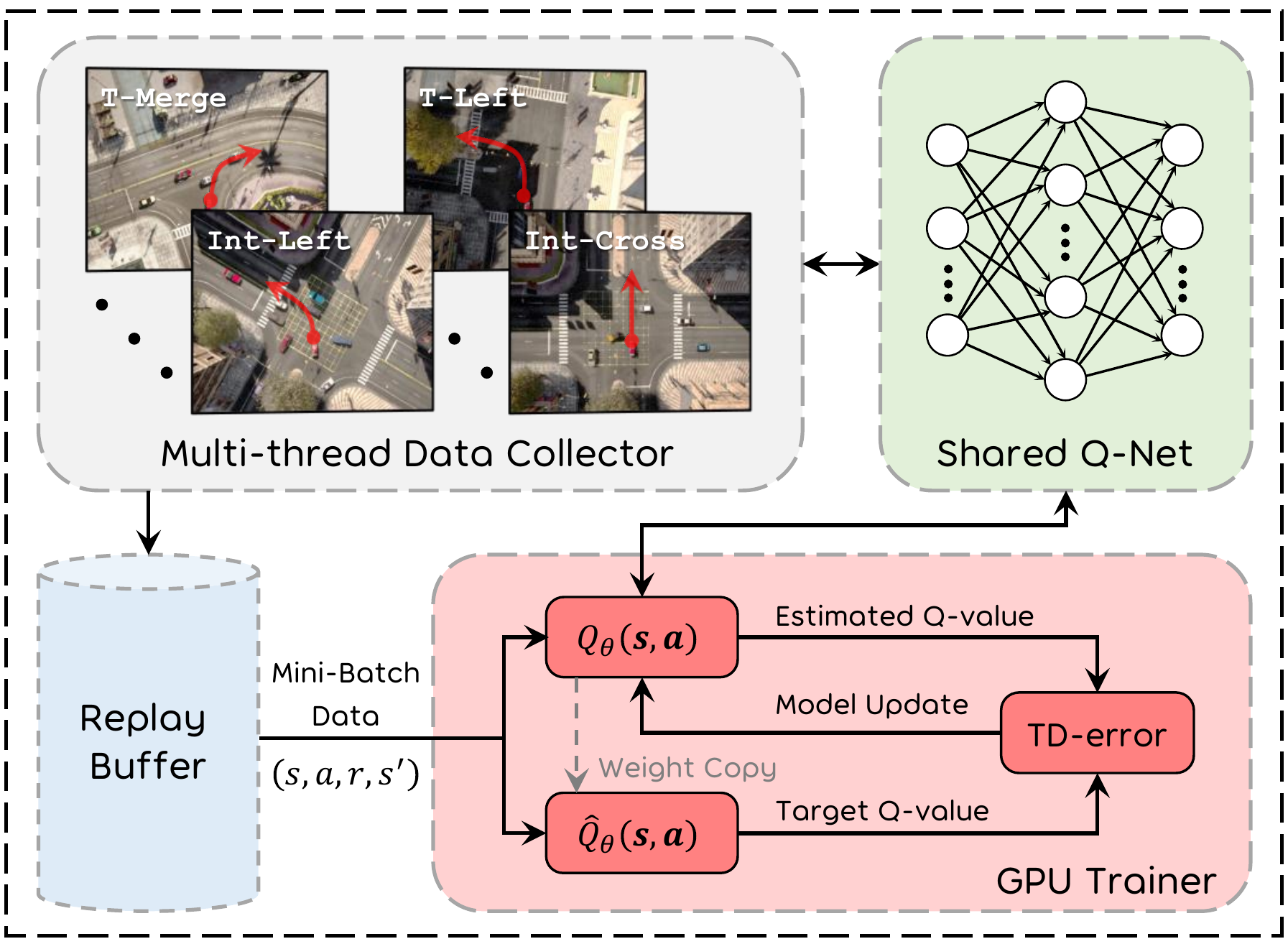}
        \caption{Overview of our training system design. The training process is divided into two iterative parts until convergence: \textit{exploration} and \textit{model update}. At the exploration stage, multiple agents sharing the same Q-network collect experiences in their separate CARLA threads. Different threads are synchronized using the message passing interface (MPI), and their transition tuples $(s, a, r, s^{\prime})$ are pushed into the replay buffer $\mathcal{D}$ in parallel, enabling high throughput. At the training stage, mini-batches of transition data are sampled from $\mathcal{D}$ to calculate the TD-loss (Eq. \ref{eq:td-loss}) to update the Q-network.}
        \label{fig:training}
        \vspace{-0.4cm}
\end{figure}

Our system is trained and evaluated in the open-source CARLA simulator (v0.9.12)\cite{dosovitskiy2017carla} since it possesses abundant vehicle models and maps close to the real world. We choose four different \textit{unsignalized} traffic scenarios, where the agent has to drive safely and efficiently according to the intent of other vehicles, being neither too conservative nor too aggressive. These scenarios are shown in Fig. \ref{fig:training}. The \texttt{T-Merge} scenario involves making a right turn for lane merging at a T-shaped junction. The \texttt{T-Left} and \texttt{Int-Left} scenario involves making an unprotected left turn at a T-shaped junction and a four-lane intersection, respectively. \texttt{Int-Cross} involves driving straight to cross the four-lane intersection. 

We use the AI engine of CARLA to form realistic and dynamic traffic flows with several random properties, in terms of vehicle types (cars, big trucks, ambulance, etc.), destinations, densities and speeds. This randomness provides a large state space to explore, which can produce generic driving policies and avoid overfitting the policy to a specific case. The simulation step is set to 0.1 seconds, meaning the control frequency is 10 Hz for all our experiments.

\subsection{Semantic Abstraction of Driving Scenes}
\label{subsec:bev}
In order to learn good driving policies, we use semantic BEV images as the representation of driving scenes to reduce the dimensionality and redundency of raw sensory data. Furthermore, with such representation, there is no domain difference between the simulation and real world, thus the policy transfer problem\cite{Tai2017VirtualtorealDR} can be alleviated. Specifically, we rasterize different semantic elements (e.g., lane marking and obstacles) into RGB channels to form a concise and informative scene representation. As shown in Fig. \ref{fig:architecture}, our BEV input is composed of the following two parts: 1) \textit{High-definition (HD) map}:
The HD map contains the drivable area, lane markings, and the route to follow. Leveraging map information to learn driving policies is very helpful because it provides valuable structural priors on the motion of surrounding road agents. For example, vehicles normally drive on lanes rather than on sidewalks, and they should not cross solid lane markings; 2) \textit{Road vehicles}: We render the ego and other vehicles on the HD map to provide more spatial information.

In this work, our region of interest is $W$=70 {m} wide (35 {m} to each side of the ego-vehicle) and $H$=50 {m} long (35 {m} in front and 15 {m} behind the ego-vehicle). The image resolution is set to 0.25 m/pixel, which finally results in a binary BEV input $\mathbf{X}$ of size 200$\times$280$\times$3, anchored at the ego-vehicle’s current position. We use the CNN backbone ResNet-18 to project $\mathbf{X}$ into a lower-dimensional vector $\bm{z} \in \mathbb{R}^{512}$ for further operation.

\subsection{Graph Modeling of Driving Scenes}
\subsubsection{Network Architecture}
As shown in Fig. \ref{fig:architecture}, we use a GAT to model the interaction among road agents during driving, and it is composed of multiple graph layers. The input to the $i$-th layer is a set of node features, $\{\bm{h}^i_1,  \bm{h}^i_2, ...,  \bm{h}^i_N\}$, $ \bm{h}^i_k \in \mathbb{R}^{F^i}$, where $N$ is the number of nodes (agents, including the ego-vehicle), and $F^i$ is the dimensions of features in each node. Then, the information of each node $k$ is propagated to the neighboring nodes $\mathcal{N}_k$ and is used to update the node features via a self-attention mechanism, which produces the output of the layer:
\begin{equation}
\bm{h}^{i+1}_{k}=\sigma(\sum_{j \in \mathcal{N}_{k}} \alpha_{kj}(\bm{h}_{k}^{i},\ \bm{h}_{j}^{i}) \mathbf{W} \bm{h}_{j}^{i}),
\label{eq:graoh attention}
\end{equation}
where $\sigma(\cdot)$ is the ReLU activation function, $\mathbf{W} \in \mathbb{R}^{F^{i+1}\times F^{i}}$ is a shared weight matrix to be applied to each node for expressive feature transformation, $\alpha_{kj}(\cdot,\cdot)$ means the importance of node $j$ to node $k$, which is the normalized attention coefficients computed with shared weight vector $\vec{\mathbf{a}} \in \mathbb{R}^{2 F^{i+1}}$:
\begin{equation}
\alpha_{kj} (\bm{h}_{k}^{i},\ \bm{h}_{j}^{i} ) = \frac{\text{exp} ( \sigma ( \vec{\mathbf{a}}^T [\mathbf{W}\bm h^i_k||\mathbf{W}\bm h^i_j  ] ) )}{\sum_{m\in \mathcal{N}_k} \text{exp}(\sigma (\vec{\mathbf{a}}^T [\mathbf{W} \bm h_k^i || \mathbf{W} \bm h_m^i   ]  ) )},
\end{equation}
where $||$ represents the concatenation operation. Furthermore, we follow the \textit{multi-head attention} method in \cite{velickovic2018graph} to stabilize the learning process. Specifically, $S^i$ independent graph networks execute the transformation of (\ref{eq:graoh attention}) and their features are concatenated to produce the output of the i-th layer:
\begin{equation}
\bm{h}_{k}^{i+1}=\mathop{\|}\limits_{s=1}^{S^i} \sigma(\sum_{j \in \mathcal{N}_{k}} \alpha_{kj}^s(\bm{h}_{k}^{i},\ \bm{h}_{j}^{i}) \mathbf{W}^s \bm{h}_{j}^{i}).
\end{equation}

For the final layer, we employ \textit{averaging} among multiple heads rather than concatenation.

\subsubsection{Implementation Details}
In this work, we adopt a two-layer GAT and set $S^1, S^2=4, \ F^1,F^2=256$. The input features $\mathcal{C}$ include motion state information for each node (road agent) in the ego-vehicle's local coordinates. For node $k \in \{1,2, ... ,N\}$, the input feature is a 10-dimensional vector:
\begin{equation}
    \bm{c}_k=\left\{x, y, d, \psi, vx, vy, ax, ay, w, l\right\},
\end{equation}
which includes its location ($x, y$), distance to the ego-vehicle ($d$), yaw angle ($\psi$), velocity ($vx, vy$), acceleration ($ax, ay$) and size (width $w$ and length $l$). Inspired by \cite{Chen2020RobotNI}, we first pass each node state $\bm{c}_k \in \mathcal{C}$ through a multilayer perceptron (MLP) to produce a feature vector $\bm{e}_k \in \mathbb{R}^{128}$ for sufficient expressive power. For context-aware graph modeling, we then concatenate $\bm{e}_k$ with $\bm{z}$ derived from the CNN module introduced in Section \ref{subsec:bev} to generate the mixed vector $\bm{g}_k$. Then, the set $\{\bm{g}_k\}$ is sent to the GAT to output the final aggregated feature $\bm{h}^o_k  \in \mathbb{R}^{256}$, which represents the internal interactions on each node $k$. We are interested in the result $\bm{h}^o_1$ of the first node, which represents the influence on the ego-vehicle.

\subsection{Driving Policy Training with D3QN}

With the components defined above, the derived feature vector $\bm{h}_1^o$ is processed with two MLPs to generate the advantage functions and the state value, separately (as shown in Fig. \ref{fig:architecture}). Finally, the Q values are computed based on Eq. (\ref{eq:dueling}). During deployment, the action with the highest Q value will be executed to control the vehicle. In the following, we introduce the implementation details of this part.

\subsubsection{Action Space}
In this work, the agent chooses among a set of target speeds in the action space $\mathcal{A} = \{0,10,20,30,40\}$ (\textit{km/h}) to navigate the vehicle longitudinally. The chosen target speed is translated into throttle and brake actions based on a low-level PID controller. The steering control is implemented with another PID to track the route.
\subsubsection{Reward Design}
To enable \textit{safe} autonomous driving, the reward is set to -50 as punishment for collision events, and to ${v}/{40} \in [0,1]$ elsewhere to stimulate driving \textit{efficiency}, where $v$ is the speed of the agent vehicle in \textit{km/h}.

\begin{figure}[t]
        \centering
        \setlength{\abovecaptionskip}{-0.1pt}
        \includegraphics[width = \columnwidth]{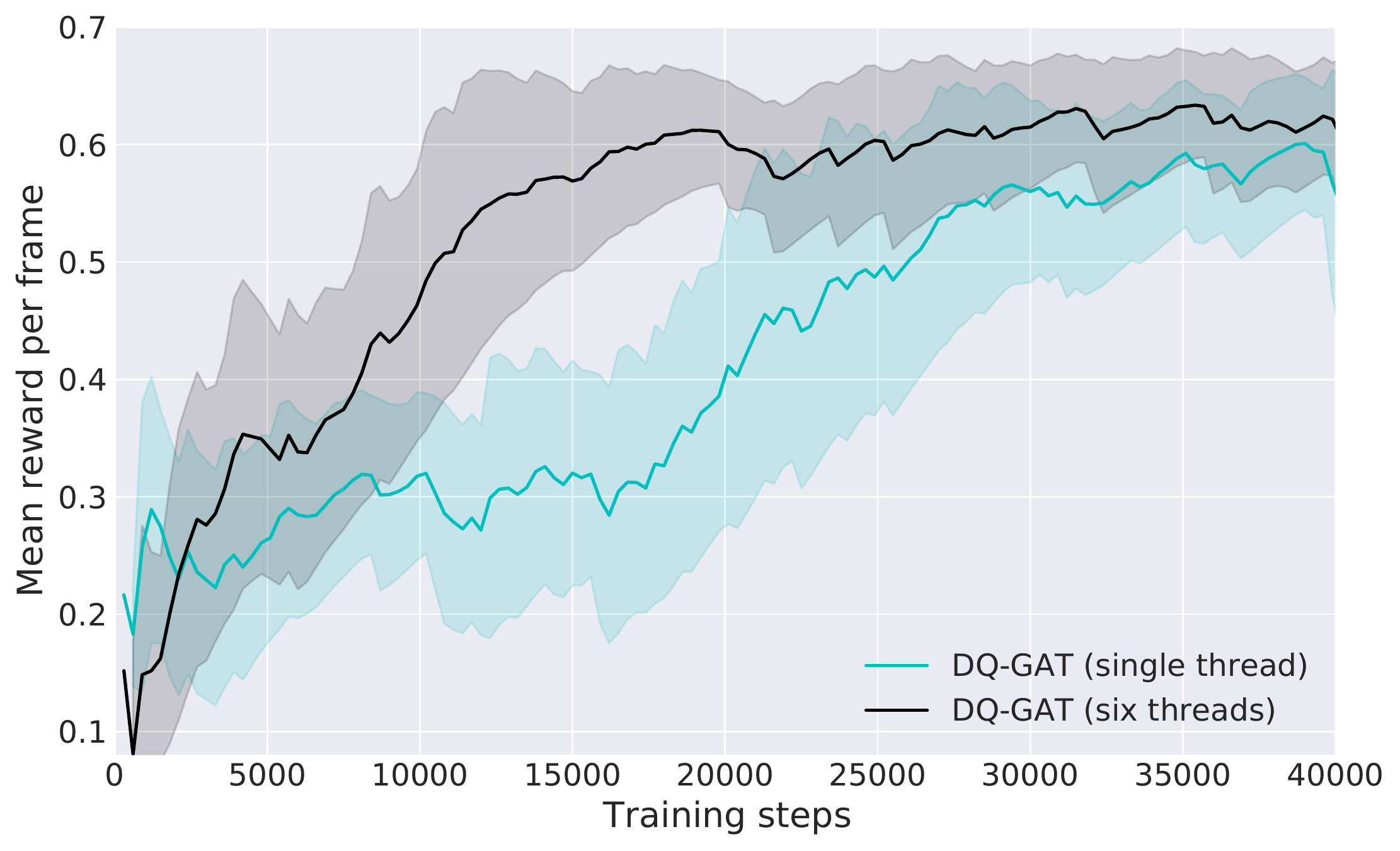}
        \caption{Training performance of our method with different numbers of experience collection threads. The solid curve indicates the mean, and the shaded region corresponds to the standard deviation.}
        \label{fig:single_thread}
        \vspace{-0.4cm}
\end{figure}

\subsubsection{Asynchronous Training} Inspired by \cite{Tai2017VirtualtorealDR}, we extend the original D3QN algorithm to an \textit{asynchronous} version with many experience collection threads working in parallel. Each thread randomly chooses a scenario to simulate for every new episode. In this way, the experience generation is decoupled from the parameter learning, which can provide higher throughput and thus improve the training efficiency. Moreover, interacting with different environments simultaneously also decorrelates the agent's data and makes the training process more stable\cite{mnih2016asynchronous}. To show the effectiveness of asynchronous training, we track the agent's mean reward during training using different numbers of threads. The results are shown in Fig. \ref{fig:single_thread}. It can be seen that the asynchronous DQ-GAT with 6 threads converges much faster than the single-thread version.

\subsubsection{Overall Pipeline} The overall training pipeline of our DQ-GAT is shown in Fig. \ref{fig:training}. The learning process is divided into two iterative parts: data collection and model update. During data collection, the agent selects and executes the actions according to the estimated Q values from $Q(s,a;\theta)$, where the noisy MLP layers are activated for exploration. Related experiences $e_{t}=\left(s_{t}, a_{t}, r_{t}, s_{t+1}\right)$ are accumulated into the buffer $\mathcal{D}=\left\{e_{1}, e_{2}, \ldots, e_{t}\right\}$. During training, we first sample mini-batches of experiences from $\mathcal{D}$ using prioritized experience replay (PER) \cite{schaul2015prioritized}, then update the parameters of the Q network through stochastic gradient descent.

\section{Experiments}
\subsection{Training Setup}
The replay buffer size is set to 500K. At each new episode, the ego-vehicle is placed in a random training scenario. The GPU trainer samples mini-batches of experiences every 4000 steps to update the Q-net for 300 rounds with the Adam optimizer. We further use a grid search to empirically find the best hyperparameters on learning rate $\alpha\in\{0.001, 0.0001\}$, reward discount $\gamma\in\{0.9, 0.99\}$ and batch size $N\in\{32, 128\}$. We found that setting $\alpha=0.0001$, $\gamma=0.99$ and $N=128$ performs best in this work.

\begin{table*}[t]
\newcommand{\tabincell}[2]{\begin{tabular}{@{}#1@{}}#2\end{tabular}}
\newcommand{\NA}{---}
        \definecolor{minigray}{rgb}{0.92, 0.92, 0.92}
        \caption{Evaluation Results of Different Models in Both Training and New Scenarios. $\uparrow$ Means Larger Numbers Are Better, $\downarrow$ Means Smaller Numbers Are Better. The Bold Font Highlights the Best Results in Each Column.}
        \label{tab:evaluation}
        \centering
        \begin{tabular}{l C{0.8cm} C{0.8cm} C{0.8cm} C{0.8cm} C{0.8cm} C{0.8cm} C{0.8cm} C{0.8cm} C{0.8cm} C{0.8cm} C{0.8cm} C{0.8cm}}
        
        \toprule
        &
        \multicolumn{8}{c}{Training Scenarios} &
        \multicolumn{4}{c}{New Scenarios}\\
        \cmidrule(lr){2-9} \cmidrule(lr){10-13}
        Scenarios&
        \multicolumn{2}{c}{\texttt{T-Left}} &
        \multicolumn{2}{c}{\texttt{T-Merge}} & 
        \multicolumn{2}{c}{\texttt{Int-Cross}} & 
        \multicolumn{2}{c}{\texttt{Int-Left}} &
        \multicolumn{2}{c}{\texttt{Five-Way}} &
        \multicolumn{2}{c}{\texttt{Roundabout}} \\
        \cmidrule(lr){2-3} \cmidrule(lr){4-5} \cmidrule(lr){6-7} \cmidrule(lr){8-9} \cmidrule(lr){10-11} \cmidrule(lr){12-13}
        Models & S.R.   & C.T.  & S.R.   & C.T. & S.R.   & C.T. & S.R.   & C.T. & S.R.   & C.T. & S.R.   & C.T.  \\
        \midrule
        {\textit{regular traffic}} & (\%) $\uparrow$ & ($s$) $\downarrow$ & (\%) $\uparrow$ & ($s$) $\downarrow$ & (\%) $\uparrow$ & ($s$) $\downarrow$ & (\%) $\uparrow$ & ($s$) $\downarrow$ & (\%) $\uparrow$ & ($s$) $\downarrow$ & (\%) $\uparrow$ & ($s$) $\downarrow$ \\
        \midrule
        
        IL-Conser. & 84.00 & 9.64& 96.00 & 8.62& 78.33 & 8.49& 83.33 & 9.27& 80.00 & 6.89& 55.67 & 30.52\\
        IL-Aggr. & 75.00 & 6.64& 84.67 & 5.31& 76.67 & 5.88& 79.33 & 6.45& 83.00 & 5.50& 26.67 & 22.79\\
        H-REIL\cite{cao2020reinforcement} & 79.67 & 6.77& 94.33 & 5.48& 81.67 & \textbf{5.83}& 82.33 & 6.53& 85.00 & 5.52& 28.33 & 22.16\\
        FSM-TTC\cite{cosgun2017towards} & 93.33 & 8.15& 99.67 & 6.48& 98.33 & 7.13& 97.00 & 7.40& 98.67 & 7.12& \textbf{87.67} & 25.66\\
        DQ-GAT\textit{(ours)} & \textbf{99.00} & \textbf{6.34}& \textbf{100.00} & \textbf{4.99}& \textbf{100.00} & 5.84& \textbf{98.33} & \textbf{6.15}& \textbf{99.67} & \textbf{5.40}& 87.33 & \textbf{21.15}\\
        
        \midrule
        {\textit{dense traffic}} \\
        \midrule
        
        IL-Conser. & 77.00 & 11.12& 92.67 & 11.99& 69.33 & 8.40& 77.67 & 11.16& 65.33 & 7.59& 21.00 & 35.57\\
        IL-Aggr. & 51.33 & 6.73& 79.00 & 5.36& 68.67 & 6.09& 66.00 & \textbf{6.49}& 67.67 & \textbf{5.39}& 2.00 & 25.32\\
        H-REIL\cite{cao2020reinforcement} & 70.00 & 6.85& 91.33 & 5.71& 69.00 & \textbf{6.02}& 76.00 & 6.54& 81.33 & 5.63& 1.00 & \textbf{22.23}\\
        FSM-TTC\cite{cosgun2017towards} & 84.33 & 10.59& 99.33 & 9.03& 94.33 & 8.48& 89.67 & 8.50& 92.33 & 10.92& 57.00 & 41.34\\
        DQ-GAT\textit{(ours)}  & \textbf{96.67} & \textbf{6.72}& \textbf{99.67} & \textbf{5.24}& \textbf{99.00} & 6.22& \textbf{98.33} & 6.63& \textbf{99.67} & 5.93& \textbf{77.67} & 31.57\\

        \bottomrule
        \end{tabular}
\end{table*}

\subsection{Training Performance with Different DRL Methods}

We first compare our DQ-GAT with several other DRL-based methods for autonomous driving to verify the effectiveness of our model design.

\begin{itemize}

    \item \textbf{GCN(U)}. This adopts a two-layer GCN to process the node features $\{\bm{g}_k\}$. We refer to the baseline \textit{U-GCNRL} introduced in \cite{Chen2020RobotNI} and set the adjacency matrix of GCN with uniform weights.

    \item \textbf{GCN(D)}. It is similar to {GCN(U)} but uses distance-related weights in its adjacency matrix. It adopts a straightforward intuition that obstacles closer to the ego-vehicle should exert a stronger influence. This network follows the idea of \textit{D-GCNRL} introduced in \cite{Chen2020RobotNI}. 
    
    \item \textbf{DenseBEV}. Following \cite{Saxena2020DrivingID}, we use the occupancy-grid style BEV image to form the observation, including information about the dynamic states of neighbouring vehicles and road layouts. Then, a CNN backbone Resnet-18 is used to handle such input.
    
\end{itemize}

\begin{figure}[t]
        \centering
        \setlength{\abovecaptionskip}{-0.1pt}
        \includegraphics[width = \columnwidth]{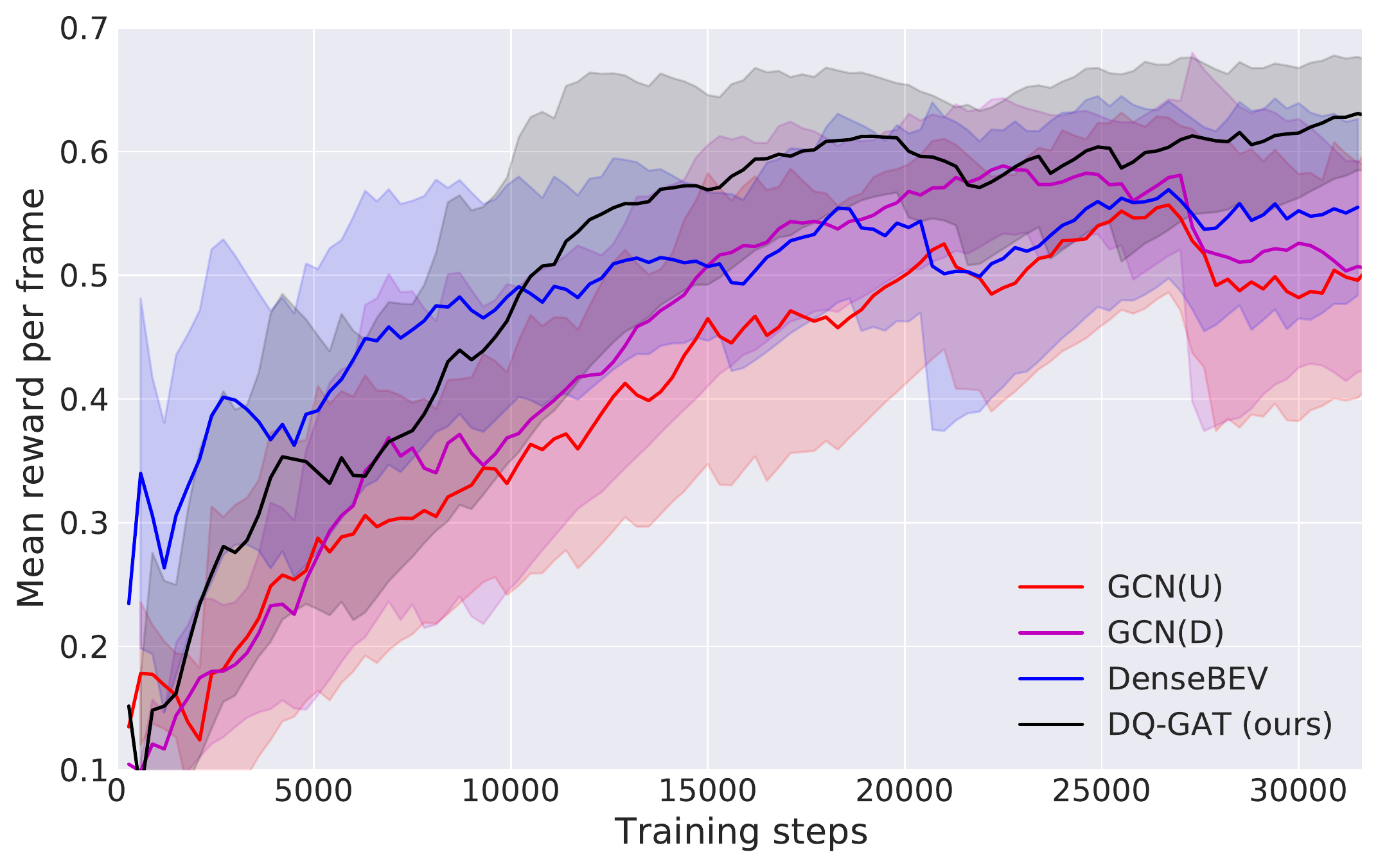}
        \caption{Training performance with different DRL methods. The solid curve indicates the mean, and the shaded region means the standard deviation.}
        \label{fig:rl-comp}
        \vspace{-0.4cm}
\end{figure}

The results are shown in Fig. \ref{fig:rl-comp}. It can be seen that our method performs better than others with higher final rewards. First, DenseBEV has information loss when rasterizing the dynamic states of vehicles into pixels. Second, the influences of road agents are not always equal (GCN(U)) or related to distance (GCN(D)). Our model addresses these problems through a heterogeneous graph attention network, where inter-vehicle influences are automatically learned and adjusted through data.

\subsection{Evaluation Methods}
\label{subsec:eval_method}
In order to cover as many driving scenarios as possible to thoroughly evaluate different methods, we set two levels of traffic densities, namely \textit{regular} and \textit{dense}. Then, we run 300 trials on each scenario setup for each model with 200 random seeds \textit{different from those in the training stage.}

\subsubsection{Metrics} We use the following two metrics, which are averaged over all episodes, to measure the driving performance: (1) S.R. (success rate): An episode is considered to be successful if the agent reaches a certain goal without any collision. The episode will be recounted if there is a traffic jam. (2) C.T. (completion time): The average time cost for the \textit{successful} trials. The \textit{failed} trials where collision happens are not counted for this metric.

\begin{figure}[t]
        \centering
        \setlength{\abovecaptionskip}{0cm}
        \includegraphics[width = \columnwidth]{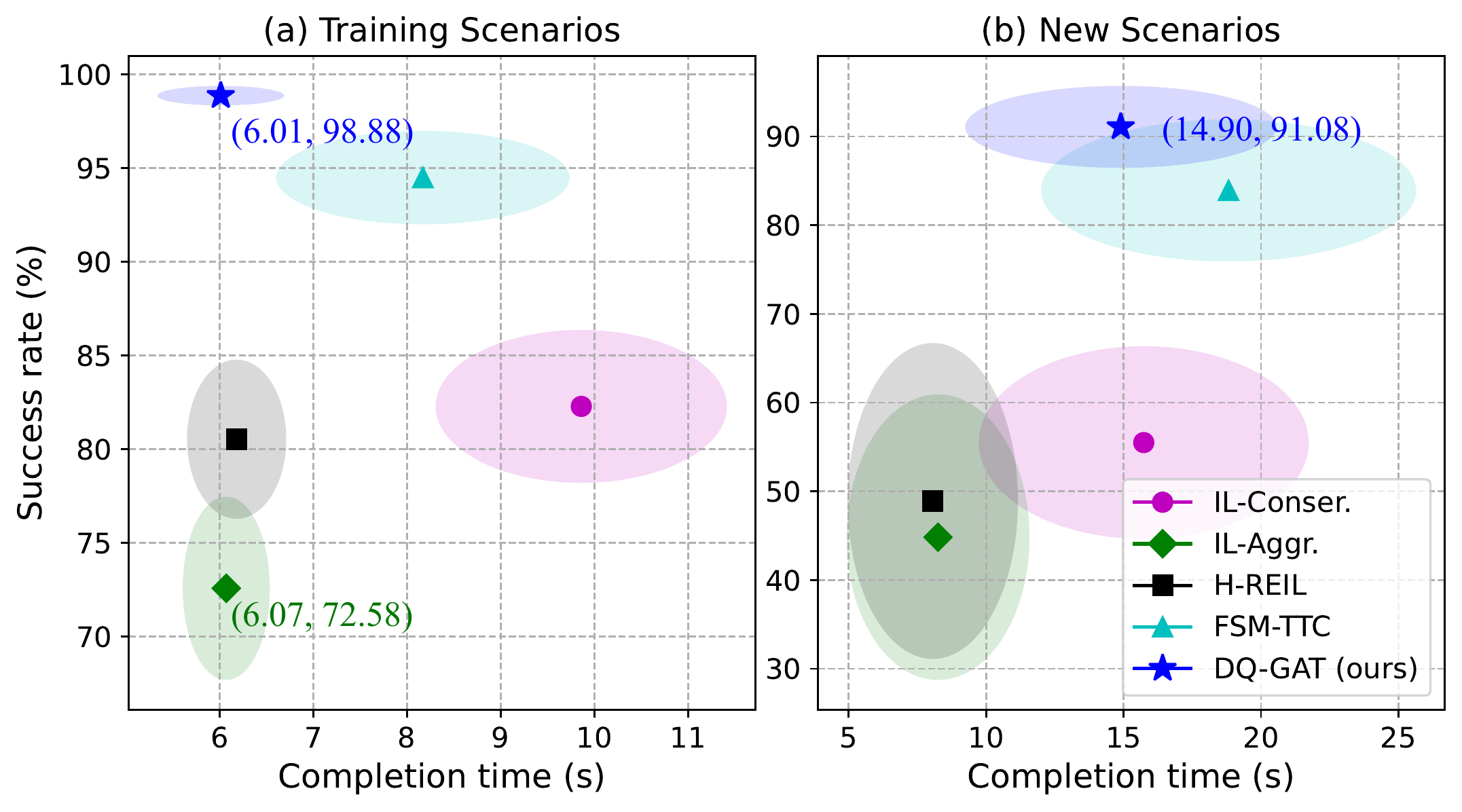}
        \caption{The average success rate and the completion time for different methods on the (a) training scenarios and (b) unseen new scenarios. The solid marker indicates the mean value and the shaded area means the standard deviation.}
        \label{fig:comp-success}
        \vspace{-0.4cm}
\end{figure}

\begin{figure*}[t]
        \centering
        \setlength{\abovecaptionskip}{-0.1pt}
        \includegraphics[width = 2\columnwidth]{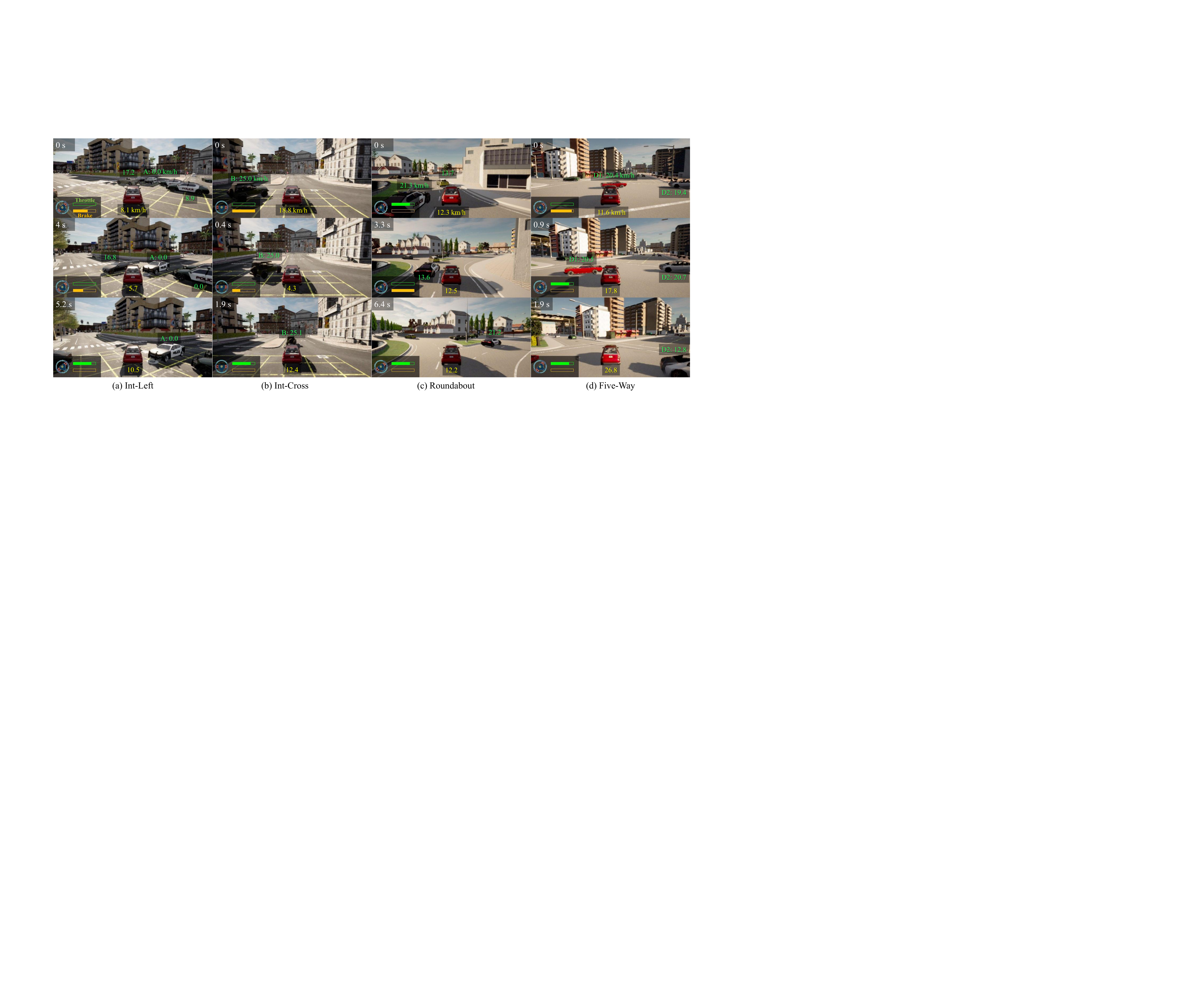}
        \caption{Evaluation results of our DQ-GAT in the CARLA simulator. We show several driving clips with camera images in four scenarios, where \texttt{Roundabout} and \texttt{Five-Way} are unseen scenarios for the agent. We label the speed of key vehicles, and render the output control commands of the ego-vehicle for better understanding. The sample driving behaviors are: (a) and (d) creeping forward to safely and efficiently drive through the traffic when taking unprotected left turns at intersections; (b) timely slowing down to avoid an accident when an aggressive vehicle beside suddenly cuts into the lane of the ego-vehicle; and (c) yielding to the vehicle (which is departing the roundabout) on the left side for collision avoidance.
        }
        
        \label{fig:qualitative}
\end{figure*}

\begin{figure}[t]
        \centering
        \setlength{\abovecaptionskip}{-0.1pt}
        \includegraphics[width = \columnwidth]{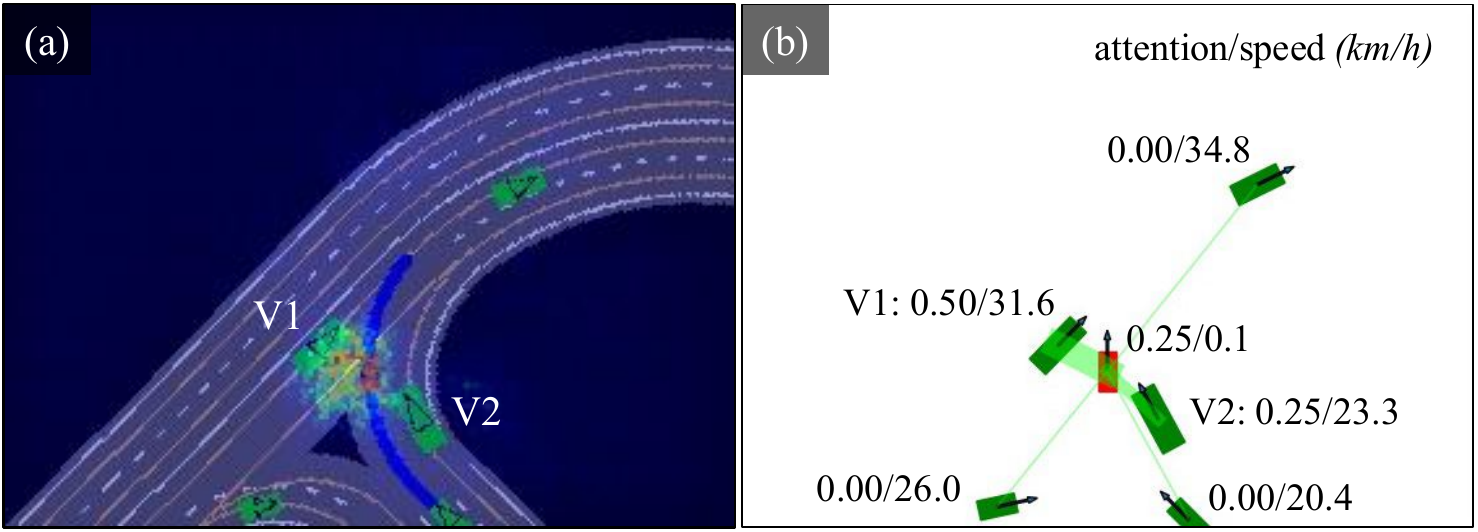}
        \caption{Policy visualization. (a) Saliency map of the input BEV image $\mathbf{X}$, where the computed Jacobian of Q is masked on the $\mathbf{X}$ for better visualization. Lighter pixels indicate more salient parts with larger values. (b) Attention distribution, where thicker lines indicate larger attentions. 
        }
        \label{fig:attention}
        \vspace{-0.4cm}
\end{figure}

\subsubsection{Baselines} 
We consider two driving modes in this paper, namely aggressive and conservative. The former favors efficiency over safety. For example, it drives fast (40 km/h) so as to reach the goal in minimal time, but tends to collide with other vehicles. By contrast, the conservative mode drives the car very cautiously (20$\sim$30 km/h), and slows down or stops to avoid all potential accidents. Accordingly, we collect human driving data in the training scenarios per mode, leading to demonstrations $\mathcal{H}_{aggr}$ and $\mathcal{H}_{conser}$, respectively. Then, we compare DQ-GAT with the following policies:
\begin{itemize}
    \item \textbf{IL-Conser}. An IL agent trained on $\mathcal{H}_{conser}$.
    \item \textbf{IL-Aggr}. An IL agent trained on $\mathcal{H}_{aggr}$.
    \item \textbf{H-REIL}. We follow \cite{cao2020reinforcement} and train a high-level DRL-based mode switcher, which selects the low-level agent, i.e., IL-Conser. or IL-Aggr., every 0.5 s to balance safety and efficiency.
    \item \textbf{FSM-TTC}. This is a traditional rule-based method for crossing intersections\cite{cosgun2017towards}. It is implemented by an FSM with the time-to-collision (TTC) safety indicator\cite{minderhoud2001extended}.
\end{itemize}

\subsection{Quantitative Analysis}
\label{subsec:quantitative}
The left column of Table \ref{tab:evaluation} shows the evaluation results in four training scenarios. The average performance is shown in Fig. \ref{fig:comp-success}-(a). We have the following main findings:

1) Since IL-Cons. favors more safety than IL-Aggr., it achieves higher success rates at the cost of efficiency. For example, in \texttt{T-Merge} with dense traffic, IL-Cons. reaches a success rate of 92.67\%, higher than that of IL-Aggr. (79\%). However, it requires much more time for the task (11.99 v.s. 5.36 s).

2) After training a high-level DRL-based mode switcher, the H-REIL model achieves success rates close to those of IL-Cons., and completion time close to those of IL-Aggr., as shown in Fig. \ref{fig:comp-success}-(a), meaning it better trades off safety and efficiency than the two base agents.

3) The rule-based method FSM-TTC achieves much higher success rates than the above methods, but it is not as efficient as IL-Aggr. with more task completion time. This is because FSM-TTC controls the vehicle in a \textit{reactive} manner, and always waits for a gap to go.

4) As shown in Fig. \ref{fig:comp-success}-(a), on average, our DQ-GAT model not only achieves the highest success rate (98.88\%), but also achieves a comparable completion time to the IL-Aggr. model (6.01 v.s. 6.07 s). These results demonstrate that DQ-GAT can better trade off safety and efficiency than previous methods. We accredit this improvement to our DRL-based pipeline, which is an unsupervised training framework enabling \textit{end-to-end} optimization (compared with H-REIL) and self-learning (compared to IL-Coners., IL-Aggr. and FSM-TTC) of driving policies.

\begin{figure*}[t]
        \centering
        \setlength{\abovecaptionskip}{-0.1pt}
        \includegraphics[width = 2\columnwidth]{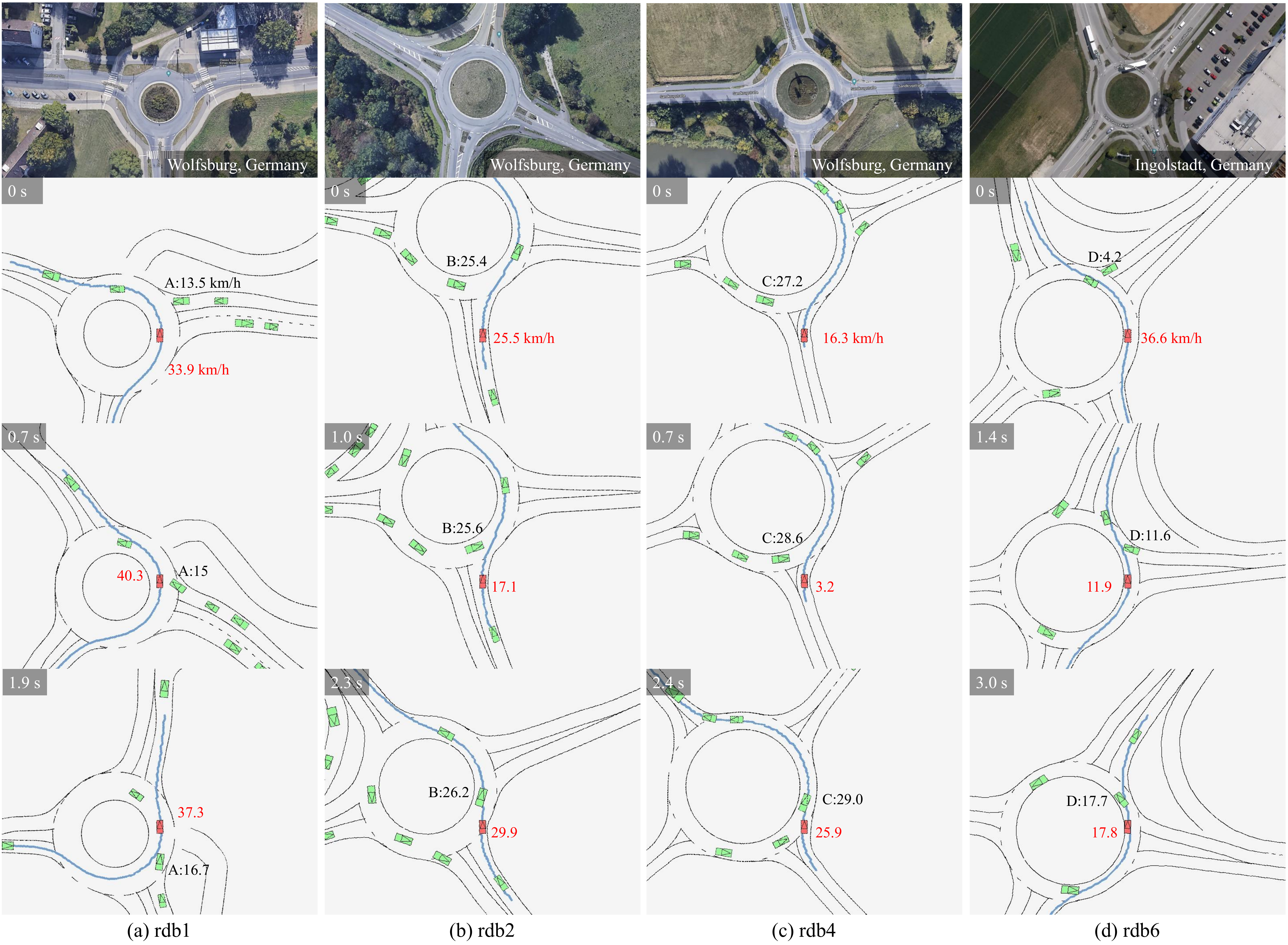}
        \caption{Qualitative results on the real-world openDD dataset. (a) Maintaining the speed to take the front way of a mild slow-driving vehicle, and decelerating to yield to the front vehicle which is (b,c) at relatively high speeds or (d) accelerating to enter the roundabout.}
        \label{fig:opendd}
        \vspace{-0.4cm}
\end{figure*}

\subsection{Qualitative Analysis}

The qualitative results of DQ-GAT in diverse dynamic environments with different road structures and traffic flows are shown in Fig. \ref{fig:qualitative}. For example, in (a) \texttt{Int-Left}, the ego-vehicle is taking an unprotected left turn at a four-lane intersection, with many other vehicles driving towards different directions in front. The ego-vehicle first applies a brake to slow down for collision avoidance at t=0 s. In the meantime, another vehicle A is also waiting to cross the intersection. Rather than stopping at the intersection and waiting other vehicles to leave, the ego-vehicle releases the brake to slowly creep forward at 5.7 km/h in an \textit{exploratory} manner (t=4 s). This interactive manner informs the other vehicles of its intent. Therefore, vehicle A continues to wait and yields to the ego-vehicle. Finally at t=5.2 s when the ego-vehicle arrives in front of vehicle A, it starts to accelerate to finish the turn efficiently.

In some cases, instant safety is more important than efficiency. For example, in (b) \texttt{Int-Cross}, when the ego-vehicle is driving straight down the road, an aggressive vehicle B suddenly cuts into the lane of the ego-vehicle at 25 km/h. To ensure safety, the ego-vehicle applies a full brake to lower the speed from 18.8 km/h to 4.3 km/h at t=0.4 s, and then accelerates at t=1.9 s to drive behind vehicle B.

In summary, these exploratory and interactive driving styles are quite similar to how humans drive, leading to a better trade-off between safety and efficiency, which can also explain the advantage of DQ-GAT on the quantitative results in Sec. \ref{subsec:quantitative}.

\subsection{Policy Visualization}
To better understand the roles of Q-learning and the attention mechanism in this work, in the following, we compute and visualize the salient part\cite{wang2016dueling} of the input BEV image $\mathbf{X}$, and show attention values from the graph networks. Specifically, to show the saliency map of $\mathbf{X}$, we compute the absolute value of the Jacobian of the estimated Q value with respect to the input BEVs: $|\nabla_{X}Q(s, \arg \max_{a^{\prime}}Q(s, a^{\prime}); \theta)|$. 

A sample driving case in the \texttt{T-Merge} scenario is shown in Fig. \ref{fig:attention}. We can see that the ego-vehicle is yielding to the front vehicle V1, which drives at high speeds (31.6 km/h) and exerts a strong influence with higher attention (0.5) than the others. Accordingly, as observed in the saliency map, the Q-value also cares more about V1. Note that there is another vehicle V2 in the right lane, which is about the same distance from the ego agent as is V1. However, V2 is assigned with lower attention (0.25), because it is turning into a different lane and does not have much influence on the ego agent. This phenomenon demonstrates that the ideas of GCN(U) and GCN(D) are not reasonable in some occasions because the influences of road agents should be measured within specific contexts and are not always equal or related to distance. By contrast, without external supervisions, our DQ-GAT can still learn to dynamically and reasonably focus on different parts of the environment, like humans do, thanks to the self-attention mechanism. Such a difference also explains the performance gap between GCN methods and our DQ-GAT in Fig. \ref{fig:rl-comp}.

\subsection{Zero-Shot Generalization Performance}
\subsubsection{New Scenarios in the CARLA Simulator}
To examine whether our method can generalize to unseen environments, we further conduct benchmark tests as stated in Sec. \ref{subsec:eval_method} in two new scenarios, namely \texttt{Roundabout} and \texttt{Five-Way}, where the vehicle should drive around a roundabout and take a left turn at a five-way intersection, respectively. The quantitative results are shown in the right column of Table \ref{tab:evaluation}. The average performance is shown in Fig. \ref{fig:comp-success}-(b). We can see that IL-based methods generalize poorly in new scenarios. For example, in \texttt{Roundabout} with regular traffic, the success rate of IL-Aggr. is only 26.67\%. By contrast, DQ-GAT achieves higher success rates (77.67$\sim$99.67\%), with similar completion time to the IL-Aggr. model, in most scenarios\footnote{An exception is in \texttt{Roundabout} with dense traffic. We observe that the IL-Aggr./H-REIL model exhibits low-level interactive driving skills, and tends to collide with other vehicles, leading to a very low success rate (1$\sim$2\%). Therefore, it can only aggressively finish a few random cases where safe behaviors that take time (e.g., yielding, vehicle-following, etc.) are not needed, leading to a shorter average completion time than ours.}.

For qualitative analysis, we demonstrate two driving cases in columns (c and d) of Fig. \ref{fig:qualitative}. In (c) \texttt{Roundabout}, the ego-vehicle timely slows down at t=3.3 s to yield to the vehicle on the left side (which is departing the roundabout) for collision avoidance. In (d) \texttt{Five-Way}, the ego-vehicle first slows down to yield to vehicle D1 for collision avoidance, then accelerates at t=0.9 s to inform vehicle D2 to slow down (20.7$\rightarrow$12.8 km/h), and finally crosses the traffic flow at t=1.9 s. In summary, similar to the performance in seen environments, DQ-GAT can still dynamically adjust its driving style in new environments according to specific driving contexts.

\subsubsection{New Scenarios in the Real World} For practical application, we further test the driving performance of DQ-GAT in openDD\cite{breuer2020opendd}, which is a real-world trajectory dataset focusing on interaction-intensive unregulated roundabouts with varying topologies in Wolfsburg and Ingolstadt, Germany. Qualitative results on four roundabouts in openDD, rdb$_1$, rdb$_2$, rdb$_4$ and rdb$_6$ are shown in Fig. \ref{fig:qualitative}. We can see that the agent makes two different decisions in these scenarios: (a) maintaining the speed against a mild vehicle for \textit{efficiency}, and (b-d) decelerating to yield to the front aggressive vehicle for \textit{safety}.

In addition, the proposed model also observes a notable inference rate of 150 Hz on the NVIDIA RTX 2060 mobile GPU, and 260 Hz on the GTX 1080 Ti GPU, essential for real-time driving applications.

\begin{figure}[t]
        \centering
        \setlength{\abovecaptionskip}{-0.1pt}
        \includegraphics[width = \columnwidth]{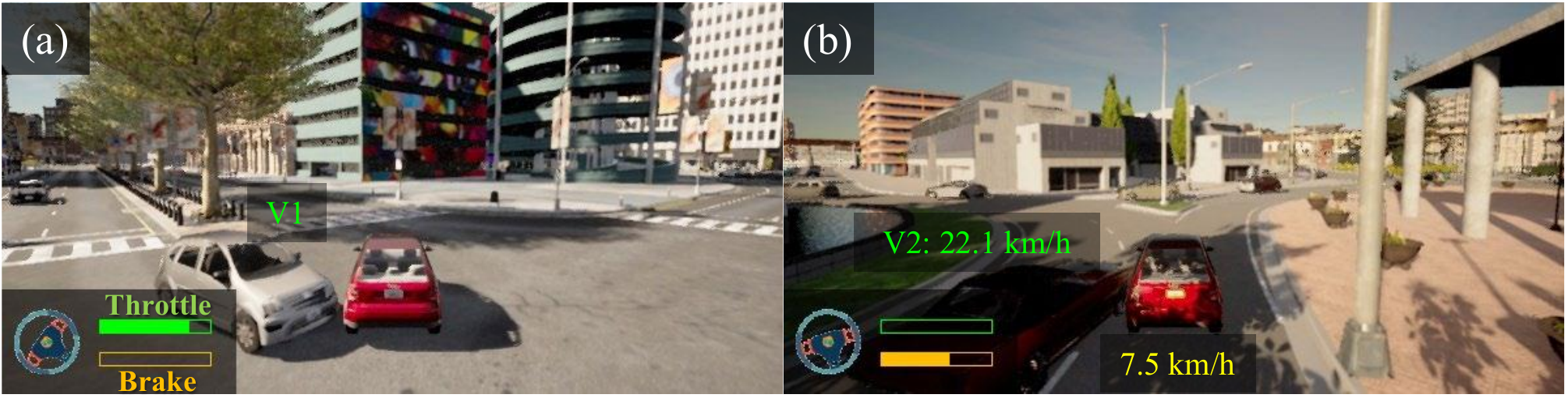}
        \caption{Sample failure cases of our method: (a) scratching accidents; (b) collisions caused by other aggressive vehicles.
        }
        \label{fig:failure}
        \vspace{-0.4cm}
\end{figure}

\subsection{Failure Cases}
\label{subsec:failure}
The typical failure cases of our model are shown in Fig. \ref{fig:failure}. We observe that they can be divided into two categories: inter-vehicle scratching accidents (Fig. \ref{fig:failure}-(a)), and collisions caused by other aggressive cars (Fig. \ref{fig:failure}-(b)). For example, in scenario (a), the ego-vehicle is taking an unprotected left turn, but its rear left panel slightly collides with the stopped vehicle V1. In scenario (b), the rear-left vehicle V2 suddenly turns right to depart the roundabout at a relative higher speed of 22.1 km/h. The ego-vehicle timely infers its intent and starts to decelerate by applying large brake values. However, V2 continues to drive without slowing down. Finally, these two vehicles collide. The limitation of our model in handling these near-accident cases leaves possible avenues for future research on safer autonomous driving.

\section{Conclusion and Future Work}

In this work, we proposed DQ-GAT to achieve safe and efficient autonomous driving in various urban environments. It is a graph-based network using the self-attention mechanism to encode heterogeneous information in generic traffic scenarios. We extended the original D3QN to an asynchronous version to train the network without relying on human labels. Then, by setting various traffic flows in different scenarios (e.g., unprotected left turns at intersections, merging, and crossing), we extensively evaluated different methods and demonstrated that our DQ-GAT can dynamically adjust its driving styles according to specific driving contexts like human drivers do, finally achieving a better trade-off between safety and efficiency than previous rule-based and learning-based methods. Afterwards, we showed that our method generalizes well in totally unseen scenarios like roundabouts and five-way intersections, where the performance of baseline learning-based methods degrades a lot in terms of safety. Furthermore, we qualitatively tested the zero-shot generalization performance of DQ-GAT, which is trained in a simulator, on the real-world dataset openDD and demonstrated its potential for practical applications.

This work makes a common assumption on perfect perception results. However, measurement noise is inevitable in the real world. In the future, we will investigate how to handle the perceptual uncertainties within the training framework.

\bibliographystyle{IEEEtran}
\bibliography{main.bib}

\begin{IEEEbiography}[{\includegraphics[width=1in,height=1.25in,clip,keepaspectratio]{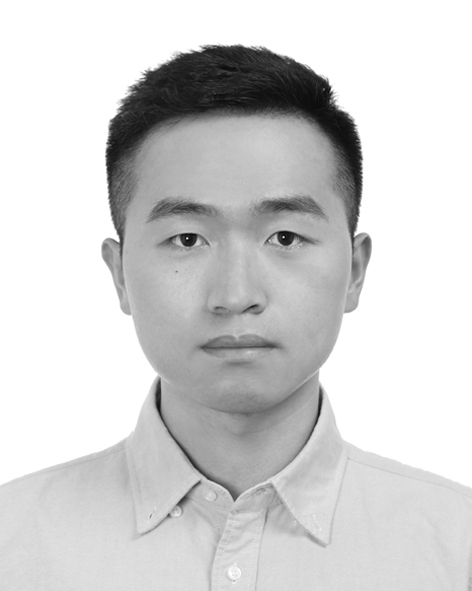}}]{Peide Cai} received the Bachelors degree of Automation in 2018 from school of Control Science and Engineering from Zhejiang University. He is currently pursuing a Ph.D. degree at the Robotics Institute, Department of Electronic and Computer Engineering, Hong Kong University of Science and Technology (HKUST), Hong Kong, China. 

His research interests include autonomous driving, robotics and autonomous systems, and deep learning.
\end{IEEEbiography}

\begin{IEEEbiography}[{\includegraphics[width=1in,height=1.25in,clip,keepaspectratio]{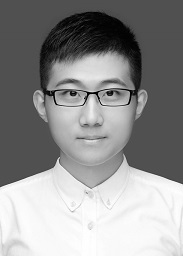}}]{Hengli Wang}
    received the B.E. degree in mechatronics engineering from Zhejiang University, Hangzhou, China, in 2018. He is now a Ph.D. student at the Robotics Institute, Department of Electronic and Computer Engineering, Hong Kong University of Science and Technology (HKUST), Hong Kong, China. 
    
    His research interests include computer vision, robot navigation and deep learning.
\end{IEEEbiography}

\begin{IEEEbiography}[{\includegraphics[width=1in,height=1.25in,clip,keepaspectratio]{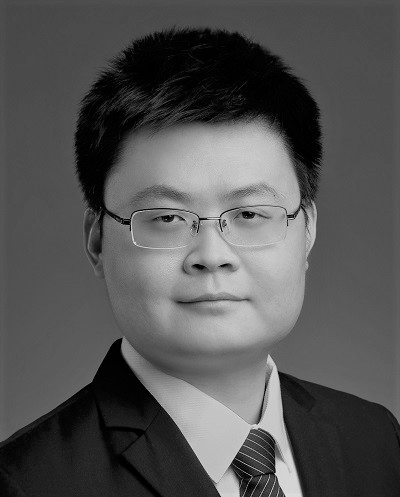}}]{Yuxiang Sun} received the Ph.D. degree from The Chinese University of Hong Kong, Hong Kong, in 2017. He is currently a Research Assistant Professor with the Department of Mechanical Engineering, The Hong Kong Polytechnic University, Hong Kong. He is an associate editor of IEEE Robotics and Automation Letters. 

His research interests include autonomous driving, robotics and artificial intelligence, deep learning, mobile robots, etc. 

\end{IEEEbiography}

\begin{IEEEbiography}[{\includegraphics[width=1in,height=1.25in,clip,keepaspectratio]{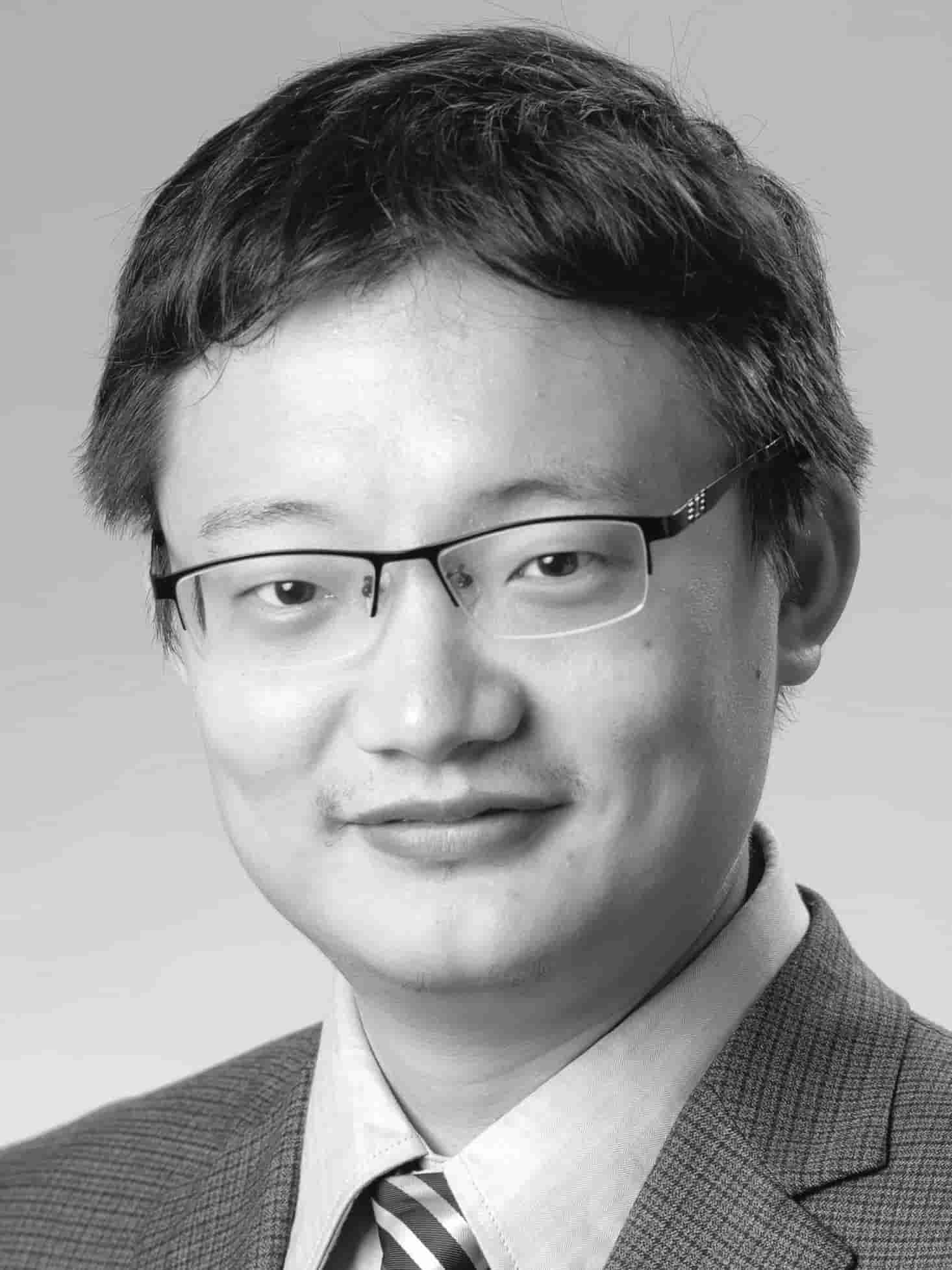}}]{Ming Liu} received the Ph.D. degree from the Department of Mechanical and Process Engineering, ETH Zürich, Zurich, Switzerland, in 2013.

He is currently an Associate Professor with the Thrust of Robotics \& Autonomous Systems, The Hong Kong University of Science and Technology (Guangzhou), China, and also with the Department of ECE, The Hong Kong University of Science and Technology, Hong Kong SAR, China. His research interests include dynamic environment modeling, 3-D mapping, machine learning, and visual control. 

\end{IEEEbiography}

\end{document}